\documentclass[letterpaper]{article} 
\usepackage{aaai2026}  
\usepackage{times}  
\usepackage{helvet}  
\usepackage{courier}  
\usepackage[hyphens]{url}  
\usepackage{graphicx} 
\usepackage{natbib}  
\usepackage{caption} 
\usepackage{algorithm}
\usepackage{algorithmic}
\usepackage{amsmath}
\usepackage{newfloat}
\usepackage{listings}
\DeclareCaptionStyle{ruled}{labelfont=normalfont,labelsep=colon,strut=off} 
\floatstyle{ruled}
\newfloat{listing}{tb}{lst}{}
\floatname{listing}{Listing}
\title{MMRAG-RFT: Two-stage Reinforcement Fine-tuning for Explainable Multi-modal Retrieval-augmented Generation}
\author{
    Shengwei Zhao\textsuperscript{\rm 1}\equalcontrib, 
    Jingwen Yao\textsuperscript{\rm 1}\equalcontrib,
    Sitong Wei\textsuperscript{\rm 1},
    Linhai Xu\textsuperscript{\rm 1},
    Yuying Liu\textsuperscript{\rm 1},
    Dong Zhang\textsuperscript{\rm 1},
    Zhiqiang Tian\textsuperscript{\rm 2},
    Shaoyi Du\textsuperscript{\rm 1}\thanks{Corresponding author.}
}
\affiliations{
    \textsuperscript{\rm 1}National Key Laboratory of Human-Machine Hybrid Augmented Intelligence, National Engineering Research Center for Visual Information and Applications, and Institute of Artificial Intelligence and Robotics, Xi'an Jiaotong University\\
    \textsuperscript{\rm 2}School of Software Engineering, Xi'an Jiaotong University\\

    \{zhaoshengwei, yaoooo, weist0422\}@stu.xjtu.edu.cn, xlh@mail.xjtu.edu.cn, yuyingliu@xjtu.edu.cn, dongzhangcv@gmail.com, \{zhiqiangtian, dushaoyi\}@xjtu.edu.cn
}

\usepackage{bibentry}

\begin{document}

\maketitle

\begin{abstract}
Multi-modal Retrieval-Augmented Generation (MMRAG) enables highly credible generation by integrating external multi-modal knowledge, thus demonstrating impressive performance in complex multi-modal scenarios. However, existing MMRAG methods fail to clarify the reasoning logic behind retrieval and response generation, which limits the explainability of the results. To address this gap, we propose to introduce reinforcement learning into multi-modal retrieval-augmented generation, enhancing the reasoning capabilities of multi-modal large language models through a two-stage reinforcement fine-tuning framework to achieve explainable multi-modal retrieval-augmented generation. Specifically, in the first stage, rule-based reinforcement fine-tuning is employed to perform coarse-grained point-wise ranking of multi-modal documents, effectively filtering out those that are significantly irrelevant. In the second stage, reasoning-based reinforcement fine-tuning is utilized to jointly optimize fine-grained list-wise ranking and answer generation, guiding multi-modal large language models to output explainable reasoning logic in the MMRAG process. Our method achieves state-of-the-art results on WebQA and MultimodalQA, two benchmark datasets for multi-modal retrieval-augmented generation, and its effectiveness is validated through comprehensive ablation experiments.
\end{abstract}

\section{Introduction}

In recent years, multi-modal large language models have attracted extensive attention due to their comprehensive understanding capabilities and powerful interaction abilities in complex multi-modal scenarios. Existing multi-modal large language models, such as GPT-4V \cite{achiam2023gpt} and LLAVA \cite{liu2024improved}, have achieved impressive performance in tasks like visual question answering, image captioning, and open-vocabulary recognition by implicitly storing world knowledge in billions of parameters of neural networks. However, similar to large language models, multi-modal large language models still face challenges in updating world knowledge and eliminating hallucinations \cite{ye2022unreliability}. As a solution, Retrieval-Augmented Generation (RAG) integrates external knowledge references to update the latest knowledge and ensure high-credibility generation \cite{guu2020retrieval,lewis2020retrieval,feng2025hyper}. Recent studies \cite{chen2022murag,yang2023enhancing,yang2023progressive,bai2025ramqa} have proposed applying Retrieval-Augmented Generation to the multi-modal domain. The question-related reference documents retrieved from external multi-modal sources are utilized to assist in generating more accurate responses. 

\begin{figure}\centering
  \includegraphics[width=0.48\textwidth,height=0.27\textwidth]{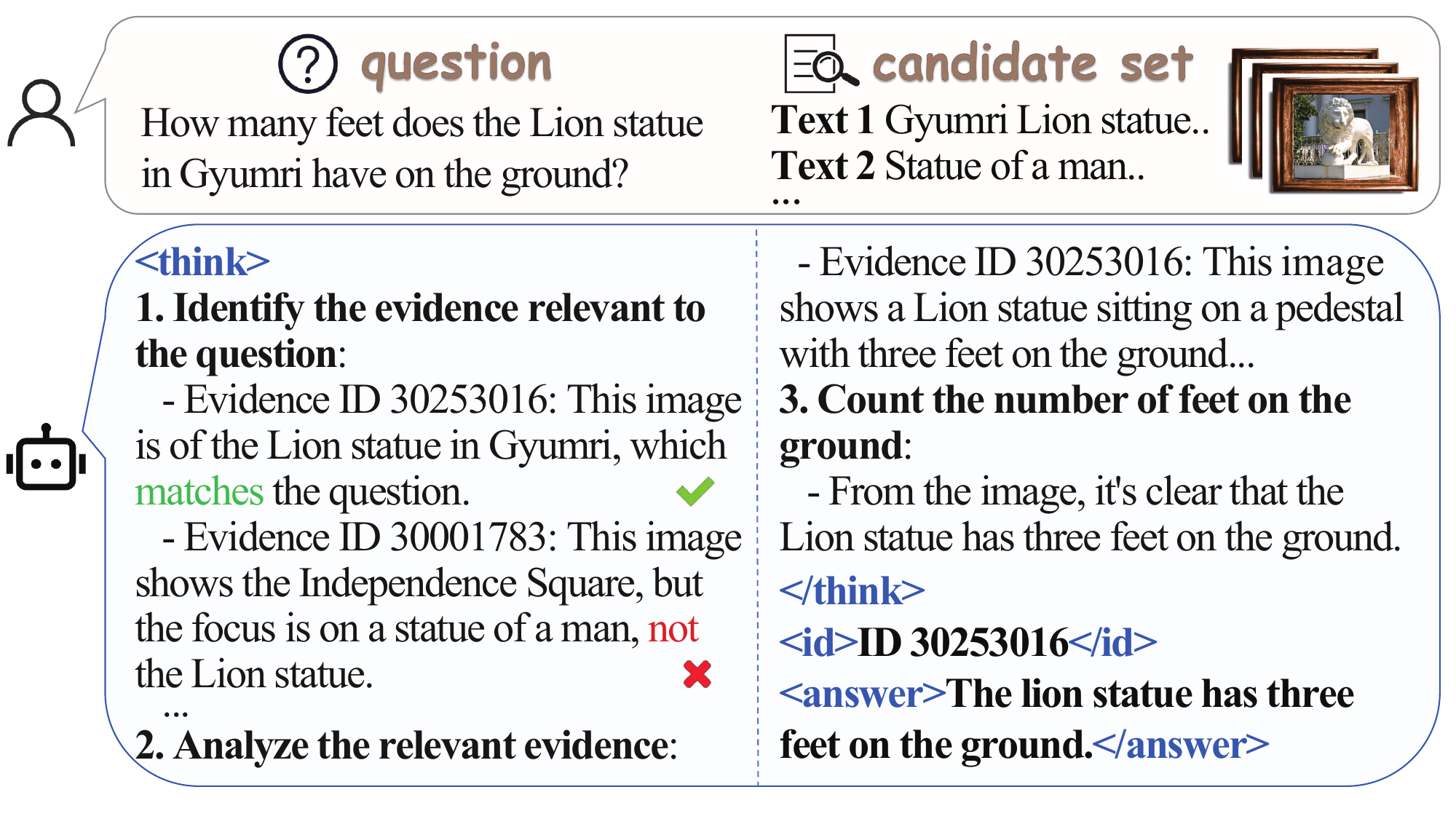}
\caption{An example of explainable multi-modal retrieval-augmented generation. The reinforcement fine-tuned model obtains the most relevant document and the final answer based on clear reasoning logic.}
\end{figure}

However, existing multi-modal retrieval-augmented generation methods (MMRAG) only stop at presenting the retrieval results and question responses, and fail to clarify the reasoning logic of retrieving specific documents and outputting corresponding responses accordingly, which is of great importance for the explainability of the generation results. To this end, a crucial question emerges: How can we conduct multi-modal explainable reasoning in the process of multi-modal retrieval-augmented generation as shown in Figure 1? To address this issue, a straightforward approach is to perform supervised fine-tuning (SFT) on multi-modal large language models based on the data with the labeled multi-modal chain-of-thought reasoning process. Nevertheless, limited by the high cost of data annotation, in the field of multi-modal retrieval-augmented generation, it is extremely difficult to obtain high-quality multi-modal chain-of-thought reasoning data. Therefore, there is an urgent need to explore the potential of multi-modal retrieval-augmented generation to develop reasoning capabilities in the absence of chain-of-thought supervision data. 

Fortunately, the latest advancements in the field of natural language processing (NLP), such as DeepSeek-R1 \cite{guo2025deepseek}, have revealed that even without supervised fine-tuning, large-scale reinforcement learning can be employed to significantly enhance the reasoning capabilities of large language models, especially in the fields of science (e.g., mathematics) and code generation. Inspired by the remarkable success of reinforcement learning in the field of NLP, We integrate reinforcement learning into multi-modal retrieval-augmented generation, where the reasoning capabilities of multi-modal large language models are enhanced via a two-stage reinforcement fine-tuning framework, thus facilitating explainable multi-modal retrieval-augmented generation. Specifically, considering the limited context length of multi-modal large models, point-wise ranking is more efficient than list-wise ranking in large-scale document ranking. In the first stage, rule-based reinforcement fine-tuning is first employed to guide the multi-modal large language model in performing coarse-grained point-wise multi-modal document ranking, thereby initially filtering out multi-modal documents that are significantly irrelevant to the query. Meanwhile, since point-wise ranking only considers the relevance score of individual documents and ignores the relationships between them, its discriminative ability is limited. Therefore, in the second stage, reasoning-based reinforcement fine-tuning is utilized to jointly optimize fine-grained list-wise ranking and the answer generation, based on the multi-modal document list after coarse-grained filtering. Among them, carefully designed preset prompts are leveraged to steer the multi-modal large language model to explore the multi-modal chain of thought, thereby explaining the justifications for fine-grained list-wise ranking and the reasoning logic for answering questions based on the most relevant documents.

Furthermore, we have noticed that existing multi-modal retrieval-augmented datasets suffer from issues such as annotation errors and uneven category distribution. Therefore, we have constructed the Mini-WebQA training dataset containing 5,000 high-quality samples. To demonstrate the effectiveness of our proposed approach, we conduct extensive experiments on multiple multi-modal retrieval-augmented generation datasets, including WebQA dataset \cite{chang2022webqa}, MultimodalQA dataset \cite{talmor2021multimodalqa}, and the Mini-WebQA dataset we constructed. Compared with previous methods, our proposed approach in multi-modal retrieval-augmented generation tasks achieves state-of-the-art results, and even yields highly competitive results when trained solely on the Mini-WebQA dataset. Our main contributions can be summarized as follows:

\begin{itemize}
\item We integrate reinforcement learning into multi-modal retrieval-augmented generation, enabling multi-modal large language models to enhance their reasoning capabilities in the absence of chain-of-thought supervision data and thereby facilitating explainable multi-modal retrieval-augmented generation.

\item We designed a two-stage reinforcement fine-tuning framework for multi-modal large language models, which first performs coarse-grained document filtering via efficient point-wise ranking, then conducts joint optimization of fine-grained list-wise ranking and answer generation, effectively balancing retrieval efficiency and response quality while enabling explainable reasoning.

\item Our method achieves state-of-the-art results on WebQA and MultimodalQA, two benchmark datasets for multi-modal retrieval-augmented generation, with comprehensive ablation experiments validating its effectiveness.  
\end{itemize}

\section{Related Work}
\subsection{Multi-Modal Retrieval-Augmented Generation}
In recent years, multi-modal retrieval-augmented generation has attracted increasing attention, as it boosts open-world question answering by incorporating relevant multi-modal information from various knowledge sources. Early methods build structured knowledge representations via encoder models to select multi-modal documents, then use questions and retrieved documents jointly for decoder generation. Examples include MuRAG \cite{chen2022murag}, which integrates multi-modal encoders for retrieval-enhanced encoding, SKURG \cite{yang2023enhancing}, which performs multi-modal multi-hop QA via entity-centric fusion encoder, and PERQA \cite{yang2023progressive}, which selects key documents through iterative retrieval.
However, these studies overlook the potential of multi-modal large models in document understanding and ranking. Recently, RAMQA \cite{bai2025ramqa} applies SFT to guide multi-modal large language models in point-wise ranking, and VLMT \cite{lim2025vlmt} constructs a multi-modal large model for retrieval-augmented generation via supervised fine-tuning. Though they perform well, both ignore reinforcement learning’s role in enhancing reasoning ability and the importance of explainable reasoning. Thus, we designed a two-stage reinforcement fine-tuning framework to unlock the potential of multi-modal large models and enable explainable reasoning.

\subsection{Large Language Model Reasoning via Reinforcement Learning}

Recent years have seen breakthroughs in using reinforcement learning (RL) to enhance the reasoning capabilities of large language models (LLMs): OpenAI-O1 \cite{jaech2024openai}, the first reasoning-focused LLM trained via large-scale RL, has achieved state-of-the-art performance across various benchmarks. Deepseek-R1-Zero \cite{guo2025deepseek} further demonstrated that pure RL frameworks can unlock the inherent reasoning potential of large language models without supervised training data, leveraging the Group Relative Policy Optimization (GRPO) algorithm \cite{shao2024deepseekmath} combined with a rule-based reward mechanism. While remarkable achievements have been made in text reasoning, the potential of reinforcement learning in multi-modal reasoning also attracts significant attention. Driven by the rapid development of the multimodal field in recent years \cite{zhao2023multi}, emerging studies \cite{shen2025vlm,liu2025visual} have extended reinforcement fine-tuning to large vision-language models, empowering them in cross-modal tasks like mathematical reasoning and open-vocabulary object detection. However, multi-modal retrieval-augmented generation based on reinforcement learning has not received specific attention in the above studies, which is precisely the core focus of our research.

\begin{figure*}\centering
  \includegraphics[width=1\textwidth,height=0.485\textwidth]{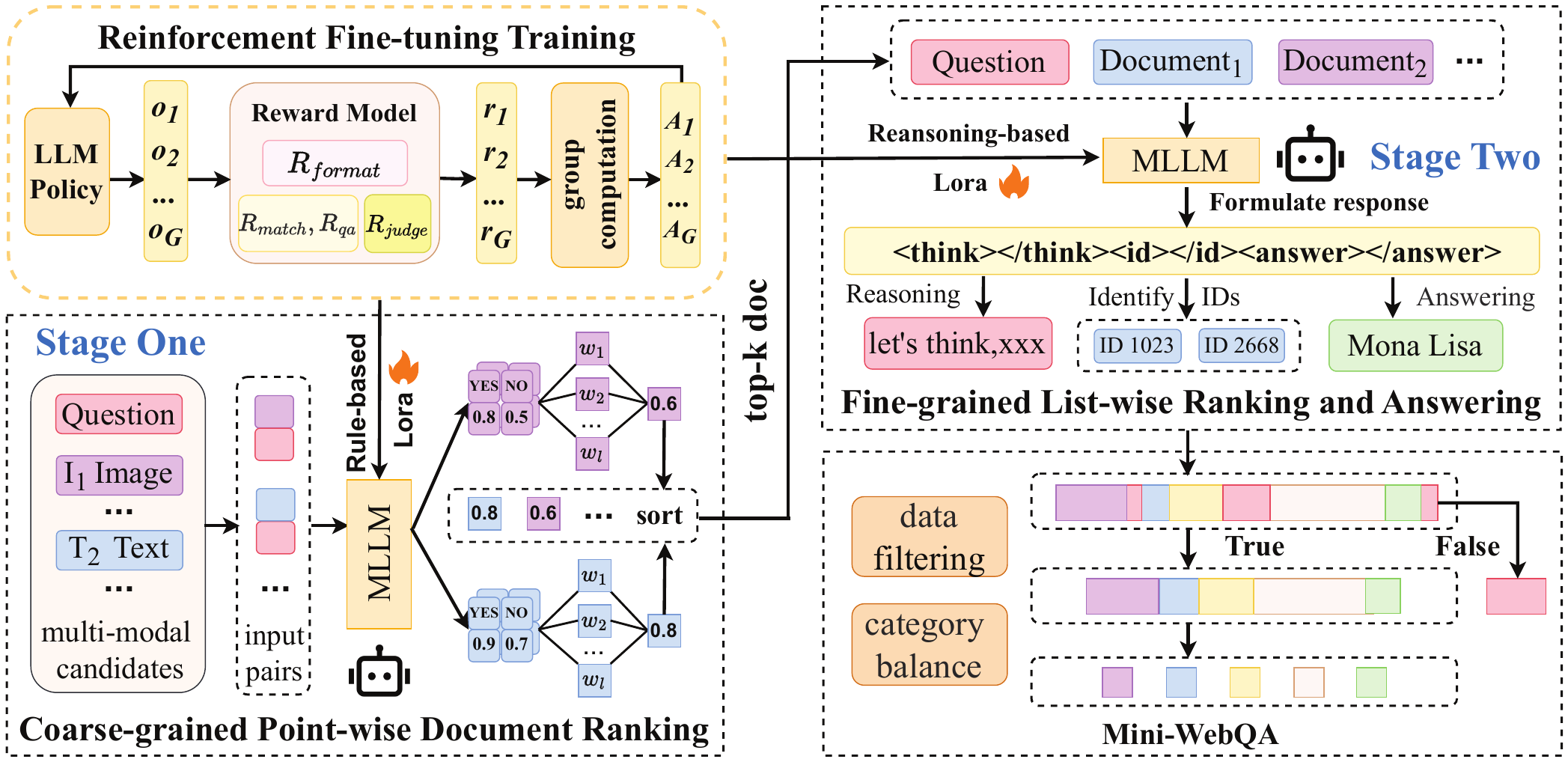}
\caption{Illustration of our proposed two-stage reinforcement fine-tuning framework.} 
\end{figure*}

\section{Methodology}
In this section, we elaborate on our two-stage reinforcement fine-tuning framework, whose overall structure is shown in Fig.2. We first introduce the problem formulation of multi-modal retrieval-augmented generation. Then, as the first stage of the framework, we specify the details of how to perform coarse-grained point-wise document ranking. Subsequently, as the second stage, we describe in detail how to jointly optimize fine-grained list-wise document ranking and answer generation based on the document list filtered by coarse-grained ranking. Finally, we present the construction process of the Mini-WebQA dataset.

\subsection{Problem Formulation}
In this work, we aim to address the task of multi-modal retrieval-augmented generation, which is to retrieve multi-modal documents relevant to the question and generate the answer based on the retrieved documents. Formally, suppose we have a set of $N$ training quadruplets, denoted as $\xi=\{(Q^q,D^q,D^t,A^t)_i\}_{i=1}^{N}$, where $Q^q$ represents the query question, $D^q=\{d_i\}_{i=1}^{n_c}$ represents $n_c$ candidate multi-modal documents, and $D^t=\{d_i\}_{i=1}^{n_t}$ represents $n_t$ target multi-modal documents related to the question, and $A^t$ represents the target answer to the question $Q^q$. Given a query question $Q^q$ and multiple candidate multi-modal documents $D^q$, our goal is to first establish a coarse-grained correlation evaluation function $f_c(Q^q,d_i)$ between the query question $Q^q$ and a candidate multi-modal document $d_i$, and retrieve the top-k multi-modal documents $D^k=\{d_i\}_{i=1}^{k}$ related to the query question based on the point-wise ranking method $S_{pw}(Q^q,D^q,f_c)$. Subsequently, we seek to jointly optimize the fine-grained list-wise multi-modal document ranking method $S_{lw}(Q^q,D^k)$ and the answer generation method $G(Q^q,D^k)$ to ensure that the generative model can give accurate answers based on optimal multi-modal reference documents. This can be formally expressed as:$$S_{lw}(Q^q,S_{pw}(Q^q,D^q,f_c))\to D^t$$ $$G(Q^q,S_{lw}(Q^q,S_{pw}(Q^q,D^q,f_c)))\to A^t$$

\subsection{Coarse-Grained Point-Wise Document Ranking}

In the first stage, given the high efficiency of point-wise ranking in large-scale document ranking, coarse-grained point-wise document ranking is employed to obtain comparable relevance scores between the query and each multi-modal candidate document, thereby preliminarily filtering out significantly irrelevant documents. To this end, we explored the point-wise ranking method for multi-modal documents based on multi-modal large language models, which encompasses two core components: rule-based reinforcement fine-tuning and point-wise probabilistic reasoning.

\subsubsection{Rule-Based Reinforcement Fine-Tuning}
The rule-based reinforcement fine-tuning aims to endow the multi-modal large language model with stronger semantic understanding ability, enabling it to effectively distinguish the degree of relevance between documents in different modalities and the query question, and thus optimize the effect of coarse-grained point-wise multi-modal document ranking. First, the initial training data $\xi=\{(Q^q,D^q,D^t,A^t)_i\}_{i=1}^{N}$ is processed into the required format for reinforcement fine-tuning. Specifically, the target multi-modal documents $\{d_i\}_{i=1}^{n_t}$ are taken as positive samples and their labels are set to "Yes", while the complement of the target multi-modal documents in the candidate documents is taken as negative samples $\{d_i\}_{i=1}^{n_c} \setminus \{d_i\}_{i=1}^{n_t}$ and their labels are set to "No". The sample in the negative sample set with the largest semantic similarity to the query question is defined as a hard negative sample $d_{neg}$. Collectively, the positive and hard negative samples $\{(Q^q_i,d_1),\cdots,(Q^q_i,d_{n_t}),(Q^q_i,d_{neg})\}_{i=1}^{N}$ are input into the multi-modal large language model as the overall query $\{x_k\}_{k=1}^{N\times(n_t+1)}$. During training, the overall query is combined with the corresponding "Yes/No" judgments $\{J_k\}_{k=1}^{N\times(n_t+1)}$ to jointly optimize the ability of the multi-modal large language model to discern semantic boundaries in relevance judgment.

Then, rule-based group relative strategy optimization algorithm is used to guide the multi-modal large language model to correct the model parameters according to the training data. Specifically, for each training sample input $x_k$, we first sample $G$ different responses $\{o_{j}\}_{j=1}^{G}$ based on the old model policy $\pi_{\theta old}$. Correspondingly, each response will obtain a corresponding reward value $\{r_j\}_{j=1}^{G}$ based on the format reward function $R_{format}(o_j)$ and the relevance judgment reward function $R_{judge}(o_j)$. The rule-based reward evaluation method is defined as follows:
$$R_{format}(o_j)=\begin{cases}
1, & \text{if } o_j =\text{Yes or } o_j=\text{No}, \\
0, & \text{otherwise}.
\end{cases}$$
$$R_{judge}(o_j)=\begin{cases}
1, & \text{if } o_j =J_k, \\
0, & \text{otherwise}.
\end{cases}$$
$$r_j=R_{format}(o_j)+R_{judge}(o_j)$$
Therefore, the reward value $r_j$ corresponding to each response $o_j$ can be obtained. By calculating the reward mean and variance of $G$ different responses $\{o_j\}_{j=1}^{G}$, the relative advantage $A_j$ of the current response can be defined as:
$$A_j=\frac{r_j-mean(\{r_1,r_2,...,r_G\})}{std(\{r_1,r_2,...,r_G\})} $$
Based on these relative advantages, we aim to optimize the policy of current model $\pi_{\theta}$. It updates the model parameters $\theta$ by maximizing the expected cumulative reward. The objective function $\mathcal{J}^{coarse}(\theta)$ for coarse-grained point-wise ranking optimization is formulated as follows:
$$\mathcal{J}^{coarse}(\theta)=\operatorname{E}[x\sim P(X),\{o_j\}_{j=1}^G\sim\pi_{\theta_{old}}(O|x)]$$
$$\frac{1}{G}\sum_{j=1}^G(\min(p_{j,\theta}A_j,clip(p_{j,\theta},1+\epsilon,1-\epsilon)A_j))$$
$$p_{j,\theta}=\pi_\theta(o_j|x)/{\pi_{\theta_{old}}(o_j|x)}$$
where $\epsilon$ is a clipping-related hyper-parameter introduced in proximal policy optimization for stabilizing training.
\subsubsection{Point-Wise Probabilistic Inference}
After rule-based reinforcement fine-tuning, point-wise probabilistic inference is used for testing, which aims to estimate the similarity between a candidate multi-modal document $d_i^{test}$ and the query question $Q^{test}$ based on the answer (Yes/No) of the multi-modal large language model. Specifically, the test query $Q^{test}$ and each document $d^{test}_i$ in the list of candidate multi-modal documents $D^{test}=\{d_i^{test}\}_{i=1}^{n_{test}}$ are reconstructed as a test samples $\{Q^{test},d_i^{test}\}$ for input into the multi-modal large language model. The probability $pro_s(Yes/No|Q^{test},d_i^{test})$ of "Yes" or "No" generated by the multi-modal large language model is used to evaluate the similarity $Sim_s(Q^{test},d_i^{test})$ between the test query and a candidate document. At the same time, considering the randomness of the large model output, $L$ responses are generated based on the current test input sampling, and the adaptive probability aggregation strategy is used to comprehensively evaluate the similarity $Sim_s(Q^{test},d_i^{test})$ between the test query and a given document. Thus, we established the point-wise relevance evaluation function $f_c(Q^{test},d_i^{test})$ between the query question and the multi-modal document, which can be defined as follows:
$$logit(Yes)=exp(pro_s(Yes|Q^{test},d_i^{test}))$$
$$logit(No)=exp(pro_s(No|Q^{test},d_i^{test}))$$
$$ Sim_s(Q^{test},d_i^{test})=\frac{logit(Yes)}{logit(Yes)+logit(No)}$$
$$w_s=\frac{exp(Sim_s(Q^{test},d_i^{test}))}{\sum_{s=1}^{L}exp(Sim_s(Q^{test},d_i^{test}))}$$
$$f_c(Q^{test},d_i^{test})=\sum\limits_{s=1}^{L}w_sSim_s(Q^{test},d_i^{test})$$
Accordingly, the point-wise relevance evaluation function is used for coarse-grained multi-modal document ranking. Specifically, the correlation score $f_c(Q^{test},d_i^{test})$ between the test query $Q^{test}$ and each document $d_i^{test}$ in the candidate multi-modal document list $D^{test}=\{d_i^{test}\}_{i=1}^{n_{test}}$ is used to construct the point-wise ranking method $S_{pw}$. After the scores are sorted in descending order, the top-k multi-modal documents $D^{test}_{topk}$ related to the query are obtained, which can be defined as follows:
$$
\begin{aligned}
S_{pw}(Q^{\text{test}},D^{\text{test}},f_c) &= argsort( f_c(Q^{\text{test}},d_1^{\text{test}}), \\
& \quad  f_c(Q^{\text{test}},d_2^{\text{test}}),\ldots,f_c(Q^{\text{test}},d_{n_{\text{test}}}^{\text{test}}) )
\end{aligned}
$$
$$ D^{test}_{topk}=D^{test}[S_{pw}(Q^{test},D^{test})[:,topk]]$$

\subsection{Fine-Grained List-Wise Document Ranking and Answer Generation}

In the second stage, addressing the limited discriminative power of point-wise ranking and the poor explainability of retrieval-augmented generation, we apply reasoning-based reinforcement fine-tuning to jointly optimize fine-grained list-wise ranking and answer generation. Based on the coarsely filtered document list, the multi-modal large language model after reinforcement fine-tuning retrieves relevant document IDs and corresponding answers to the query through explainable reasoning logic. The method consists of two core components: reasoning-based reinforcement fine-tuning and explainable chain-of-thought inference.

\subsubsection{Reasoning-Based Reinforcement Fine-Tuning}
The reasoning-based reinforcement fine-tuning aims to endow the multi-modal large language model with long-form reasoning thought chain capabilities, enabling it to generate structured semantic analyses by integrating query questions and top-k relevant documents from point-wise ranking, and thus jointly optimize the accuracy of fine-grained list-wise multi-modal document ranking and answer generation. Accordingly, similar to point-wise ranking, the training samples $\{(Q^q,D^q_{topk})_i\}_{i=1}^{N}$ are input into the multi-modal large language model as the overall query $\{x_i\}_{i=1}^{N}$. And $D^q_{topk}$ represents top-k multi-modal relevant documents obtained from the coarse-grained point-wise ranking. Meanwhile, the reasoning-based group relative strategy optimization algorithm is employed for the joint optimization of fine-grained ranking and answer generation by sampling $G$ different answers $\{o_{j}\}_{j=1}^{G}$. The rewards of the reasoning-based group relative policy optimization algorithm include format reward $R_{format}$, fine-grained list-wise ranking reward $R_{match}$ and answer generation reward $R_{qa}$. First, the format reward aims to guide the multi-modal large language model to output responses with predefined structures after performing thought chain reasoning, which can be defined as follows:$$pat_1=<think>.*?</think>$$
$$pat_2=<id>.*?</id>$$
$$pat_3=<answer>.*?</answer>$$ 
$$R_{format}(o_j)=\begin{cases}
1, & \text{if } pat_i \text{ in } o_j, i=1,2,3  \\
0, & \text{otherwise}.
\end{cases}$$
Therefrom, the thinking and reasoning process of multi-modal large language models is enclosed within the \verb!<think></think>! tags, the answer of the most relevant document ID is enclosed within the \verb!<id></id>! tags, and the question answer is enclosed within the \verb!<answer></answer>! tags. Next, the fine-grained list-wise ranking reward is designed to guide the multi-modal large language model to directly give the IDs of the most relevant documents to the query based on the query and the top-k relevant documents obtained from the point-wise ranking, which can be defined as follows:
$$R_{match}(o_j)=\frac{|ID^{list}\cap ID^{target}|}{2n_{list}}+\frac{|ID^{list}\cap ID^{target}|}{2n_t}$$
Among them, $ID^{list}=\{id_t\}_{t=1}^{n_{list}}$ represents the most relevant document IDs generated by the model between the \verb!<id></id>! tags of $o_j$, $ID^{target}=\{id_t\}_{t=1}^{n_t}$ represents IDs of target multi-modal documents related to the question, and $|ID^{list} \cap ID^{target}|$ represents the number of elements in the intersection of the sets $ID^{list}$ and $ID^{target}$. Finally, answer generation reward is intended to steer the multi-modal large language model to generate accurate answers by leveraging the most relevant documents derived from list-wise ranking, which can be defined as follows:
$$R_{qa}(o_j)=\frac{exp(Bartscore({A_i^t},{o_j}))}{exp(Bartscore({A_i^t},{A_i^t}))}$$
where $Bartscore$ represents an evaluation metric that comprehensively assesses the factuality and fluency of answers \cite{yuan2021bartscore}. Accordingly, each response $o_j$ will obtain a corresponding reward value $r_j$ based on the overall reward function, which can be defined as follows:
$$r_j=R_{format}(o_j)+R_{match}(o_j)+R_{qa}(o_j)$$
Similar to the coarse-grained point-wise ranking in the first stage, the objective function $\mathcal{J}^{fine}(\theta)$ for fine-grained list-wise ranking and answering optimization can be defined as:
$$\mathcal{J}^{fine}(\theta)=\operatorname{E}[x\sim P(X),\{o_i\}_{i=1}^G\sim\pi_{\theta_{old}}(O|x)]$$
$$\frac{1}{G}\sum_{i=1}^G(\min(p_{i,\theta}A_i,clip(p_{i,\theta},1+\epsilon,1-\epsilon)A_i))$$
\subsubsection{Explainable Chain-of-Thought Inference}
After reasoning-based reinforcement fine-tuning, explainable thought chain reasoning is used for testing, aiming to guide the multi-modal large model to generate the most relevant document ID and question answer through explainable reasoning logic. Specifically, the test query $Q^{test}$ and the top-k multi-modal documents $D^{test}_{topk}$ obtained by point-wise probability inference are jointly input into the multi-modal large language model after reasoning-based reinforcement fine-tuning. Under the guidance of preset prompts, the multi-modal large language model first performs multi-step chain reasoning between the \verb!<think></think>! tags to evaluate the semantic relevance between the query and each candidate document, and then generates a ranked list of document IDs $S_{lw}(Q^{test},S_{pw})$ between the \verb!<id></id>! tags to identify the documents most relevant to the query question, and finally generates accurate answers $G(Q^{test},S_{lw})$ between the \verb!<answer></answer>! tags based on the query and its most relevant documents. Therefore, we established the fine-grained list multi-modal document ranking method $S_{lw}$ and the answer generation method $G$, which are defined as follows:
$$
\begin{aligned}
S_{lw}(Q^{test},S_{pw}(Q^{test},D^{test},f_c)) &= \\
MLLM_{<id>}(Q^{test},D^{test}_{topk})\to D^t
\end{aligned}
$$
$$
\begin{aligned}
G(Q^{test},S_{lw}(Q^{test},D^{test},f_c)) &= \\
MLLM_{<ans>}(Q^{test},D^{test}_{topk})\to A^t
\end{aligned}
$$

\begin{table*}\centering
\renewcommand{\arraystretch}{1.1}
\begin{tabular}{l|c|cccc}
\hline
\cline{1-6}
Model & Type & Retr & QA-FL & QA-Acc & QA  \\
\hline
VLP \cite{zhou2020unified}& w/o   LLM/MLLM & 68.9 & 42.6 & 36.7 & 22.6   \\
VLP + VinVL \cite{chang2022webqa}& w/o   LLM/MLLM & 70.9 & 44.2 & 38.9 & 24.1  \\
MuRAG \cite{chen2022murag} & w/o   LLM/MLLM & 74.6 & 55.7 & 54.6 & 36.1  \\
SKURG \cite{yang2023enhancing} &  w/o   LLM/MLLM & 88.2 & 55.4 & 57.1 & 37.7  \\
PERQA \cite{yang2023progressive} &  w/o   LLM/MLLM & \textbf{89.6} & 61.7 & 63.9 & 44.4  \\
VLMT \cite{lim2025vlmt} & MLLM-Based & 87.8 &64.0& 66.7 &47.6 \\
RAMQA \cite{bai2025ramqa}& MLLM-Based & 88.4 & 64.1 & 66.6 & 48.1  \\
AETGA \cite{zhang2024entailment} &  LLM-Based & 88.6 & \underline{68.3} & \underline{72.8} & \underline{54.1}  \\
\hline
\textbf{Our Method (Trained on Mini-WebQA)} & MLLM-Based & 78.6 & 67.3 & 71.1 & 53.8  \\ 
\textbf{Our Method} & MLLM-Based & \underline{89.1} & \textbf{70.8} & \textbf{76.3} & \textbf{58.3}  \\ 
\hline
\end{tabular}
\caption{WebQA official test-set results indicated on the leaderboard. We achieve the highest result on QA-FL, QA-ACC, QA score. Bold numbers denote the best scores, and underlined numbers indicate the second-best scores.}\label{tab1}
\end{table*}

\subsection{Mini-WebQA}
During the experiment, we noticed that the existing multi-modal retrieval-augmented generation datasets have problems with annotation errors and uneven category distribution. Therefore, we set out to construct the Mini-WebQA training  dataset, which contains 5,000 high-quality samples. To eliminate low-quality samples from the training data of the WebQA dataset, we adopted two steps. First, we removed question-answer pairs with obvious grammatical errors or unrecognizable characters. Second, we used a model fully trained on the complete WebQA dataset to generate answers for each question, estimating sample clarity and representativeness via answer accuracy, and retaining questions with estimated correct answers to avoid negative impacts of incorrectly labeled samples. Following these steps, 15,000 samples were retained. Additionally, we noticed that the filtered samples still suffered from the category imbalance problem. Therefore, we analyzed the quantity distribution across categories such as color, shape, yes/no, number, and multiple-choice questions, and removed samples with severe category bias. After achieving category balance, 5,000 high-quality samples were retained.

\section{Experiments}

\subsection{Datasets}

\noindent\textbf{WebQA} 
The WebQA dataset \cite{chang2022webqa} is a multi-modal, multi-hop question-answering dataset designed to evaluate the capabilities of models in complex cross-modal reasoning and generation tasks. It contains 34.2k training question-answer pairs, 5k validation question-answer pairs, and 7.5k test question-answer pairs. Each question requires retrieving a variable number of relevant documents from approximately 40 multi-modal candidate documents for multi-step reasoning to obtain answers in the form of free-form sentences. The dataset employs two key evaluation metrics: the Retr score assesses the accuracy of models in retrieving relevant evidence documents, while the QA score evaluates the quality of answer generation by models. The QA score is a composite of the QA-FL score and the QA-Acc score. The former measures the fluency between the generated answer and the reference answer, and the latter evaluates the overlap of key entities between the generated answer and the reference answer. Among the various metrics used for WebQA evaluation , the QA score holds the most significance.

\noindent\textbf{MultimodalQA}
MultimodalQA \cite{talmor2021multimodalqa} is a multi-modal question-answering dataset with 16 distinct question types, designed to evaluate the multi-modal, multi-hop reasoning capabilities of models. Each query is accompanied by approximately 20 multi-modal distractors, and models must distinguish these distractors from correct document. Following the experimental setup in RAMQA \cite{bai2025ramqa}, we focus on the query subset defined as "ImageQ" related to image information. The performance metrics used for evaluation include exact match score (EM) and average F1 score, which measure the accuracy of the generated answers relative to the reference answers in terms of exact string matching and token overlap.

\subsection{Implementation Details}
We conducted experiments on two datasets: WebQA and MultimodalQA. Unlike previous methods that separate retrieval and generation into two modules, our proposed approach uses Qwen-2.5-VL-7B \cite{bai2025qwen2} as the backbone architecture throughout the entire process of multi-modal retrieval-augmented generation. We perform parameter-efficient reinforcement fine-tuning via group-relative policy optimization combined with low-rank adaptation (LoRA) \cite{hu2022lora}. And CLIP-ViT-L-14 is used offline for hard negative sample selection. During the reinforcement fine-tuning process, group-relative policy optimization samples 4 responses for each question every time, sets $\epsilon$ to 0.28 to ensure stable training, and does not use KL divergence constraints to encourage the model to explore diverse responses independently. The training configuration of LoRA includes a rank of 64, a scaling factor of 128, a dropout rate of 0.05. During the training process, to improve efficiency and stability, we adopt bfloat16 mixed-precision training, FlashAttention-2 and gradient checkpointing \cite{chen2016training} to accelerate model training. The generated sequence length is limited to 1024 tokens to avoid training collapse, the learning rate is set to 1e-5, the generation number $L$ is set to 4, and the data seed is set to 42 to ensure the reproducibility of the experiment.

\begin{table}[h]
\centering
\renewcommand{\arraystretch}{1.1}
\begin{tabular}{lcc}
\cline{1-3}
\textbf{Model} & \textbf{EM} & \textbf{F1}  \\
\cline{1-3}
Q-only \cite{lewis2019bart}  & 11.0 & 15.6  \\
AutoRouting \cite{talmor2021multimodalqa} & 37.8 & 37.8 \\
MuRAG \cite{chen2022murag} & 58.2 & 58.2  \\
PERQA \cite{yang2023progressive}  & 63.9 & 64.1  \\
RAMQA \cite{bai2025ramqa} & 67.0 & 67.0  \\
\hline
\textbf{Our Method} & \textbf{71.7} & \textbf{79.7} \\
\hline
\cline{1-3}
\end{tabular}
\caption{\label{citation-guide} Experiment results on MultimodalQA dataset.} 
\end{table}

\subsection{Main Result}
Tables 1 and 2 present a comparative analysis of the multi-modal retrieval-augmented generation results achieved by our proposed method against other leading methods on the WebQA dataset and MultimodalQA dataset. On the WebQA dataset, our method achieves a score of 0.708 in QA-FL, 0.763 in QA-Acc, and 0.583 in overall QA performance, surpassing all baselines on these metrics. Even when trained solely on the Mini-WebQA dataset, our method still yields highly competitive results, which we attribute to the strong adaptability of our proposed framework to high-quality small-scale data. On the MultimodalQA dataset, compared with the previous state-of-the-art RAMQA, our method achieves absolute improvements of 4.7\% and 12.7\% in EM score and F1 score. The outstanding performance highlights the superiority of our proposed two-stage reinforcement fine-tuning framework in multi-modal semantic understanding and high-quality answer generation. It is worth noting that LLM-Based methods generally outperform traditional methods based on structural knowledge representation in terms of the quality of generated answers, but their retrieval performance is inferior to the state-of-the-art vector retrieval-based methods. We attribute this to the inherent disadvantage of generative models in retrieval capabilities relative to discriminative models.

\subsection{Ablation Experiments}
In this section, we conduct comprehensive ablation experiments to clarify the impact of each module in our method. All ablation experiments are based on Qwen-2.5-VL-3B as the backbone architecture, which is trained on the Mini-WebQA dataset and tested on the WebQA official test set.

\noindent\textbf{Effect of Model Scale and Training Stage} As shown in Tab. 3, we explored the impact of model scales and the training stages within the two-stage reinforcement fine-tuning framework on model performance. Among them, the large-scale Qwen2.5vl-7B model achieves better results than the basic Qwen2.5vl-3B model in most metrics. This is because a larger model scale enhances the ability of the model to understand multi-modal information and perform complex reasoning. Regarding the training stages, Stage1 achieves an absolute improvement of 15.0\% in Retr, while Stage2 achieves an absolute improvement of 64.1\% in Retr and 26.5\% in QA respectively. This may be due to the fact that each stage of reinforcement fine-tuning enhances the multi-modal large language model in different aspects, and the reinforcement fine-tuning across different stages can promote each other.

\begin{table}[h]
\centering
\renewcommand{\arraystretch}{1.1}
\begin{tabular}{lcccc}
\cline{1-5}
\textbf{Model} & \textbf{Retr} & \textbf{QA-FL} & \textbf{QA-Acc} & \textbf{QA} \\
\cline{1-5}
Total w/ 7B model  & \textbf{78.6} & \textbf{67.3} & \textbf{71.1} & \textbf{53.8} \\
Total & 76.9 & 67.3 & 66.5 & 50.9 \\
Total w/o Stage1 & 61.9 & 65.6 & 64.6 & 49.5 \\
Total w/o Stage2 & 12.8 & 30.2 & 52.4 & 24.4 \\
\hline
\cline{1-5}
\end{tabular}
\caption{\label{citation-guide} Ablation experimental results for model scales and training stages. Stage 1:coarse-grained point-wise ranking, Stage 2: fine-grained list-wise ranking and answering.}
\end{table}

\begin{table}[h]
\centering
\renewcommand{\arraystretch}{1.1}
\begin{tabular}{cccccc}
\cline{1-6}
\textbf{Stage1} &\textbf{Stage2} & \textbf{Retr} & \textbf{QA-FL} & \textbf{QA-Acc} & \textbf{QA} \\
\cline{1-6}
RFT & RFT & \textbf{76.9} & \textbf{67.3} & \textbf{66.5} & \textbf{50.9} \\
SFT & RFT & 51.5 & 63.8 & 61.5 & 45.4 \\
RFT & SFT & 63.4 & 49.7 & 53.0 & 34.9 \\
SFT & SFT & 43.6 & 46.8 & 49.9 & 31.7 \\
\hline
\cline{1-5}
\end{tabular}
\caption{\label{citation-guide} Ablation experimental results for fine-tuning methods. RFT: reinforcement fine-tuning method, SFT: supervised fine-tuning method.}
\end{table}

\noindent\textbf{Effect of Fine-Tuning Method} To further investigate the impact of different fine-tuning methods on the multi-modal retrieval-augmented generative model, we carefully designed four models with distinct stage 1 and stage 2 training configurations (RFT/SFT). As shown in Tab. 4, the best performance was achieved when both stages were trained using reinforcement fine-tuning method (RFT). When Stage 1 was switched to supervised fine-tuning (SFT) while Stage 2 remained RFT, the Retr value dropped by 25.4\%, indicating that the first stage of reinforcement fine-tuning training is crucial to improving the retrieval ability of the model. Meanwhile, configuring Stage 1 as RFT and Stage 2 as SFT led to substantial declines in multiple indicators for evaluating answer quality (QA-FL, QA-ACC, QA), highlighting the critical role of reinforcement fine-tuning in ensuring the accuracy and fluency of generated answers. In addition, the model that uses supervised fine-tuning in both stages performs the worst on all metrics. In summary, the reinforcement fine-tuning method significantly outperforms the traditional supervised fine-tuning method across various experimental settings. We attribute this superiority to the fact that the supervised fine-tuning forces the model to overfit specific task patterns, whereas the reinforcement fine-tuning enables the multi-modal large language model to evolve autonomously without compromising the integrity of the pre-trained knowledge architecture.

\section{Conclusion}
In this paper, we propose integrating reinforcement learning into multi-modal retrieval-augmented generation to enhance reasoning capabilities of multi-modal large language models, thereby enabling explainable multi-modal retrieval-augmented generation. We design a two-stage reinforcement fine-tuning framework: the first stage involves coarse-grained point-wise ranking of multi-modal documents for efficient filtering, while the second conducts joint optimization of fine-grained list-wise ranking and answer generation, aiming for explainable reasoning. Extensive experiments on datasets demonstrate the effectiveness of our framework.

\section{Acknowledgments}
This work was supported by the National Natural Science Foundation of China under Grant Nos. U24A20252 and 62088102, the Key Research and Development Program of Shaanxi Province of China under Grant Nos. 2024PT-ZCK-66 and 2024CY2-GJHX-48.

\bibliography{aaai2026}

\end{document}